\documentclass[runningheads]{llncs}
\usepackage[T1]{fontenc}
\usepackage{graphicx,verbatim}
\usepackage{amsmath}
\usepackage{xcolor}
\usepackage{multicol}
\usepackage{multirow}
\usepackage{booktabs}
\usepackage{amssymb}
\usepackage{makecell} 
\usepackage{xspace}
\newcommand{\VM}[1]{\textcolor{orange}{\texttt{\textbf{VM: }}#1}}

\newcommand{\MS}[1]{\textcolor{red}{\texttt{\textbf{MS: }}#1}}

\newcommand{\method}{Mad\textsc{CLIP}\xspace}

\begin{document}
\title{MadCLIP: Few-shot Medical Anomaly Detection with CLIP}
%

\author{Mahshid Shiri\inst{1}  \and
Cigdem Beyan\inst{1} \and
Vittorio Murino\inst{1,2}}

\institute{
Department of Computer Science, University of Verona, Verona, Italy \and
AI for Good (AIGO) Research Unit, Istituto Italiano di Tecnologia, Genoa, Italy \\
\email{\{mahshid.shiri, cigdem.beyan, vittorio.murino\}@univr.it}
}

\maketitle              
\begin{abstract}
An innovative few-shot anomaly detection approach is presented, leveraging the pre-trained CLIP model for medical data, and adapting it for both image-level anomaly classification (AC) and pixel-level anomaly segmentation (AS). A dual-branch design is proposed to separately capture normal and abnormal features through learnable adapters in the CLIP vision encoder. To improve semantic alignment, learnable text prompts are employed to link visual features. 
Furthermore, SigLIP loss is applied to effectively handle the many-to-one relationship between images and unpaired text prompts, showcasing its adaptation in the medical field for the first time. Our approach is validated on multiple modalities, demonstrating superior performance over existing methods for AC and AS, in both same-dataset and cross-dataset evaluations. Unlike prior work, it does not rely on synthetic data or memory banks, and an ablation study confirms the contribution of each component. The code is available at \url{https://github.com/mahshid1998/MadCLIP}.
 

\end{abstract}
\section{Introduction}
\label{sec:introduction}

Medical anomaly detection (AD) involves identifying unusual patterns in medical data, a task complicated by the lack of a universal anomaly definition, inconsistent patterns, and noisy data from variably calibrated sensory devices. These challenges are magnified by the crucial role of AD in medical diagnosis, where high sensitivity is essential. Therefore, AD models in medicine must achieve exceptional performance for clinical reliability \cite{fernando2021deep,su2021few,zhang2020viral}.

Overall, the AD task is approached from two main perspectives in the broader literature: (a) unsupervised techniques (e.g., \cite{deng2022anomaly,salehi2021multiresolution}) and (b) supervised methods (e.g., \cite{huang2024adapting,zhang2024mediclip,yao2023explicit,wang2022medclip}). Unsupervised methods detect anomalies by leveraging large datasets of normal samples, modeling the normal data distribution, and identifying anomalies as deviations. For example, PatchCore \cite{roth2022towards} compares test samples to a memory bank of normal embeddings and measures the nearest distance, while CFLOW-AD \cite{gudovskiy2022cflow} models normal samples with a Gaussian distribution using normalizing flows.


While many methods rely on large datasets, real-world applications often include a few labeled anomalies, which provide valuable, application-specific insights and enable recent models to significantly improve the detection of similar anomalies \cite{ding2022catching}. 
In this context, supervised methods operate in a \textit{few-shot AD} setting, where both normal and anomalous samples are limited, e.g., \cite{ding2022catching,huang2022registration,sheynin2021hierarchical,yao2023explicit}. However, the limited samples available during training, for both normal and anomalous classes, often fail to capture their full variability, restricting the model's ability to generalize to unseen cases \cite{ding2022catching}.

Recently, CLIP \cite{radford2021learning} based methods have made significant strides in few-shot AD for medical images, e.g., \cite{zhang2024mediclip,huang2024adapting}. It is clear that, relying solely on CLIP \cite{radford2021learning} is insufficient, as its training focuses on aligning with the class semantics of foreground objects, limiting its ability to generalize and capture subtle visual abnormalities, and restricting its direct application in AD.
Also, the substantial distribution shift between the data on which CLIP \cite{radford2021learning} was trained and medical images results in suboptimal performance when CLIP is applied directly to medical AD \cite{huang2024adapting}.
To effectively leverage CLIP for few-shot AD, it is crucial to address the domain gap and fine-tune or adapt CLIP \cite{radford2021learning} specifically for the medical AD task. For instance, MVFA \cite{huang2024adapting} utilizes visual adapters in the form of fully connected layers, while MediCLIP \cite{zhang2024mediclip} uses convolutional layers. On the other hand, several studies have shown that leveraging text modality as a representative of normal and abnormal classes can aid AD \cite{jeong2023winclip,chen2023zero,zhang2024mediclip,huang2024adapting}. Since anomaly descriptions might share similarities across different datasets, incorporating textual information reduces reliance only on visual data and, might enhance model performance, particularly in data-scarce scenarios such as few-shot learning. 
For instance, WinCLIP \cite{jeong2023winclip} uses a large set of artificial text prompts, while April-GAN \cite{chen2023zero} maps visual features extracted from CLIP \cite{radford2021learning} onto the linear space of text features in addition to using several memory banks. In medical AD, MVFA \cite{huang2024adapting} builds on the principles of April-GAN \cite{chen2023zero} by employing multi-level adaptation of CLIP and utilizing fixed prompts. 
Unlike MVFA \cite{huang2024adapting}, WinCLIP \cite{jeong2023winclip}, and April-GAN \cite{chen2023zero} using fixed prompts, MediCLIP \cite{zhang2024mediclip} adopts the learnable prompts approach from \cite{zhou2022coop}. Learnable prompts offer a key advantage over fixed prompts, which require careful design and expert knowledge for medical scenarios \cite{zhang2024mediclip}. Besides visual feature adaptation and text prompts, another common approach is using memory banks (e.g., \cite{chen2023zero,huang2024adapting}), though this strategy is relatively costly and often fails to generalize well. Some, e.g., \cite{zhang2024mediclip}, generate extensive synthetic data for the abnormal class to improve generalization.

Our approach, \textbf{\method} extends CLIP~\cite{radford2021learning} with a two-branch architecture using adapters to capture normal and abnormal visual features. We further leverage the role of text in data-scarce scenarios, using it to represent both normal and abnormal distributions separately. \method is thus designed to learn these distributions from both visual data and text, aiming to obtain a clear distinction between normal and abnormal patterns. As a result, complementary branches exchange signals, and two sets of learnable prompts enable the model to capture distinctive descriptions for both normal and abnormal patterns. In detail, \method models normal and abnormal representations separately within a dual optimization process, maximizing multimodal (i.e., text and vision) similarity within each class while minimizing it between classes, thereby simplifying decision-making by subtracting (i.e., contrasting) learned feature representations to achieve better class separation.
This enhances AD performance, particularly in cross-dataset settings, while also our pipeline avoids the need for memory banks or additional synthetic data.
To perform image-text alignment, unlike the standard Softmax-based loss, we use SigLip \cite{zhai2023sigmoid}, justified in the next section, where we also show its performance benefits with an ablation study. \method was evaluated on six datasets across five medical modalities using both a standard and cross-dataset approach. It outperforms state-of-the-art (SOTA) methods in anomaly classification (AC) and segmentation (AS). 

The main contributions are: \textbf{(1)} A novel few-shot AD architecture with multi-level adapters, each focusing on either normal or abnormal instances, enhanced by a dual optimization objective utilizing learnable text embeddings for better separation. This approach does not require extensive synthetic data or memory banks, unlike SOTA methods. \textbf{(2)} This is the first application of SigLIP loss \cite{zhai2023sigmoid} in medical AD, proving its effectiveness. \textbf{(3)} Strong generalization and improved performance are demonstrated through extensive validation and cross-dataset evaluation across diverse medical modalities and anatomical areas.

\begin{figure}[t!]
    \centering
    \includegraphics[width=1\linewidth]{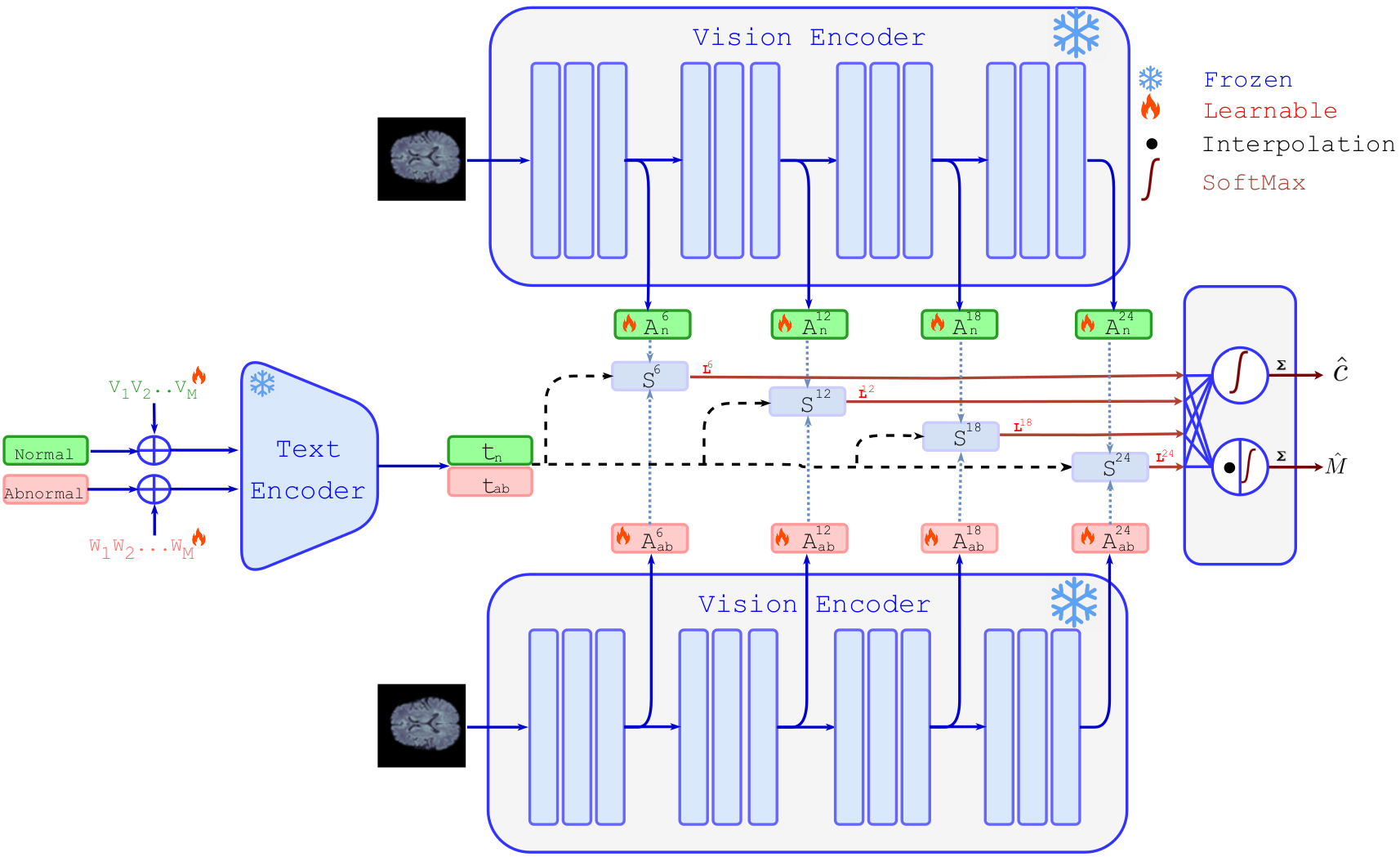}
    \caption{
    Overview of \method: A dual-branch design integrates adapters \( A_n \) and \( A_{ab} \) into CLIP’s vision encoder to separately capture normal and abnormal features. Learnable text prompts \( V_1, \dots, V_M \) and \( W_1, \dots, W_M \) encode complementary semantics for AD. The outputs are image-level AC \( \hat{c} \) and AS mask \( \hat{M} \).}
    \label{im:proposed}
\end{figure}

\section{Method}
\label{sec:method}

Few-shot medical AD is based on a dataset consisting of tuples
$\{(x, c, M)\}$, where each $x \in \mathbb{R}^{h \times w \times 3}$ represents a training image with spatial dimensions $h \times w$, $c \in \{0, 1\}$  denotes the image-level AC label ($1$ for anomalous, $0$ for normal), and, when available, $M \in \{0, 1\}^{h \times w}$ provides the pixel-level AS map. The training set is balanced, ensuring that $
|\{x \mid c = 0\}| = |\{x \mid c = 1\}|$. Given a test image $x_{\text{test}}$, the model predicts both AC and AS. 
Building on this setup, we propose \method (see Fig.~\ref{im:proposed}), a novel approach that leverages CLIP~\cite{radford2021learning}, incorporating multi-level visual features adaptation and learnable text prompts to enhance performance. \method is a dual-branch architecture that learns separate multi-modal representations for normal and abnormal samples. 
Below, we provide a detailed description of the components employed in \method. \\

\noindent \textbf{Vision Adapters.} \method employs adapters (denoted as $A^i_n$ and $A^i_{ab}$ for normal and abnormal samples, respectively) within the CLIP vision encoder, which is pre-trained on natural images~\cite{radford2021learning}, to effectively adapt it for medical imaging and the two target tasks: AC and AS.
Using adapters follows the findings of~\cite{huang2024adapting}, which shows that it is preferable to traditional fine-tuning, as it helps avoid overfitting due to high model complexity and limited data. 
Our adapters are integrated while keeping the backbone frozen. In detail, for an input image \( x \), we extract the \( i \)-th layer feature from the CLIP vision encoder, denoted as \( I^i(x) \in \mathbb{R}^{G \times d} \). Here, \( G \) represents the grid size, \( d \) is the feature dimension, and \( i \in \{6,12,18,24\} \). 
Learnable adapters consist of two transformation stages. The first stage focuses on addressing the domain gap between natural and medical images. Since CLIP embeddings are inherently optimized for object-level tasks in natural images, they may not directly align with the requirements of medical AD. To bridge this gap, we introduce a shared linear transformation layer ($ W^i_{\text{shared}}$) that refines the CLIP embeddings, mapping them into a feature space better suited for medical AD, formulated as \( F^i_{\text{shared}}(I^i(x)) = \text{ReLU}(W^i_{\text{shared}} I^i(x)) \).
The second stage, given the need to perform both AC and AS, focuses on extracting features specialized for each task, we apply two distinct linear transformation heads on top of the shared features: the first head, dedicated to AC, captures high-level characteristics essential for detecting anomalies, while the second head, tailored for AS, emphasizes fine-grained spatial details, expressed as  
\( F_{\text{Det}}^i(I^i(x)) = \text{ReLU}\Big(W^i_{\text{det}}\, F_{\text{shared}}^i(I^i(x))\Big) \)  
and  
\( F_{\text{Seg}}^i(I^i(x)) = \text{ReLU}\Big(W^i_{\text{Seg}}\, F_{\text{shared}}^i(I^i(x))\Big) \).  \\
%
%
\\ \noindent \textbf{Learnable Prompts.} Prior works have shown that well-designed text prompts in CLIP can encapsulate rich semantic information, resulting in more reliable and transferable representations for several tasks~\cite{li2024promptad}. By using learnable prompts, e.g., \cite{zhou2022coop}, we can eliminate the complexity of manually engineered prompts, which we argue that it leads to better generalization of the resulting text embeddings to medical imaging tasks. Additionally, we posit that these prompts can provide complementary signals between our dual branches, enhancing AD performance.
Standard prompts such as ``A photo of a [CLS]'' primarily capture the overall semantic content of images, which often fails to reflect the subtle, domain-specific details found in medical imaging~\cite{zhou2023anomalyclip,huang2024adapting}.
To address this limitation, 
we develop a template for the normal ($p_n$) and abnormal ($p_{ab}$) classes as: $p_n = [V_1][V_2]\ldots[V_M][\text{CLS(normal)}][\text{Objective}]$, $\quad p_{ab} = [W_1][W_2]\ldots[W_M]$ $[\text{CLS(abnormal)}][\text{Objective}]$
where \( [V_i] \) and \( [W_i] \) denote learnable token embeddings, \( [\text{CLS(normal)}] \) and \( [\text{CLS(abnormal)}] \) are fixed class embeddings, and \( [\text{Objective}] \) encodes the fixed semantic context of the target modality (e.g., Brain). 
To further enhance AD, we leverage an ensemble of text prompts by incorporating multiple synonyms for \textit{normal} (e.g. flawless, unblemished) and \textit{abnormal} (e.g. with a flaw, disease).
Each synonym generates a distinct prompt, leading to two prompt sets, $P_n = \{p_{n_1}, p_{n_2}, \dots, p_{n_k}\}\text{\xspace and \xspace}P_{ab} = \{p_{ab_1}, p_{ab_2},\dots, p_{ab_k}\}$, where \( k \) represents the number of synonyms. By aggregating the diverse textual prompts from these sets, we obtain two final prompts, \( t_n \) and \( t_{ab} \). \\

\noindent \textbf{Dual Branch Architecture.}
In few-shot AD, normal samples \( x_n \) are assumed to be drawn from an unknown distribution \( D_n \) (i.e., \( x_n \sim D_n \)), while anomalous samples \( x_{ab} \) originate from a distinct, typically unknown distribution \( D_{ab} \). These two distributions are roughly complementary such that \( D_{ab} = 1 - D_n \), based on the assumption that anomalies are defined as deviations from normal data. Our dual-branch architecture processes these distributions via two specialized branches.
The so-called \textit{normality branch} extracts features from normal samples and aligns them with a learnable text prompt \(t_n\), capturing the behavior of \(D_n\). In parallel, the so-called \textit{abnormality branch} processes abnormal samples and aligns the extracted features with a complementary prompt \(t_{ab}\). Each branch is trained to maximize the cosine similarity ($cosSIM$) between its visual features and the corresponding prompt while minimizing similarity with the opposing prompt (i.e., representative of the opposite class). 
    
Formally, for a normal sample \(x_n\), adapters \(A^i_n\) produce feature representations \( O^i_n = A^i_n\big(I^i(x_n)\big) \). 
The dual optimization objective for normal samples is to maximize 
\(\text{cosSIM}(O^i_n, t_n) - \text{cosSIM}(O^i_n, t_{ab})\),  which, assuming cosine similarity approximates the dot product, simplifies to
\(\mathbf{\max\left[O^i_n \cdot t_n - O^i_n \cdot t_{ab}\right]}\). An analogous formulation is used for abnormal samples: \(\mathbf{\max\left[O^i_{ab} \cdot t_{ab} - O^i_{ab} \cdot t_n\right]}\). These objectives aim to enforce a clear separation between the normal and abnormal feature spaces.
Furthermore, at each feature layer \(i\), the normality and abnormality scores are computed as \( S^i_n = \left[O^i_n \cdot t_n - O^i_n \cdot t_{ab}\right] \quad \text{and} \quad S^i_{ab} = \left[O^i_{ab} \cdot t_{ab} - O^i_{ab} \cdot t_n\right] \)
and then concatenated into a single vector $S^i = \left[S^i_n,\, S^i_{ab}\right]$. 
During inference, as $S$ is in patch-level, for AC we calculate the mean score over all patches, and for AS as we need to match the input size, we use bilinear interpolation to obtain image-level score vectors and aggregate them across layers
, i.e., $\hat{M}^i = \text{SoftMax}\big(\text{Interpolate}(S^i)\big), \quad \hat{c}^i = \text{Mean}\big(\text{SoftMax}(S^i)\big)$
where $\hat{M}=\frac{1}{|i|}\sum_i \hat{M}^i$ is the predicted anomaly map and $\hat{c}=\frac{1}{|i|}\sum_i \hat{c}^i$ is predicted anomaly score. Here, $|i|$ refers to the total number of feature levels at which adapters are integrated into the visual encoder. This multi-layer adaptation aims to effectively integrate complementary information from both branches for robust AD. \\

\noindent \textbf{Loss Function.}
The composite loss function we use 
at feature level $i$ is \( L^i = \lambda_1\, \text{Dice}(\hat{M}^i, M) + \lambda_2\, \text{Focal}(\hat{M}^i, M) + \lambda_3\, \text{SigLip}(\hat{c}^i,\, c) \)
where $\hat{M}^i$ represents the predicted anomaly map and  $\hat{c}^i$ is the predicted anomaly score at the $i$'th feature level, $M$ is the ground truth mask, and $c$ denotes the image-level anomaly label. The hyperparameters $\lambda_1$, $\lambda_2$, and $\lambda_3$ are fixed to 1. $Dice(.,.)$, $Focal(.,.)$, and $SigLip(.,.)$ correspond to Dice~\cite{milletari2016v}, Focal~\cite{ross2017focal}, and a sigmoid-based loss for text-image alignment~\cite{zhai2023sigmoid}. The Dice and Focal losses are necessary for AS, and particularly Focal loss is preferred as there is a significant class imbalance at the pixel level, with anomalous pixels being greatly outnumbered by normal pixels. Instead, for the AC task, given the balanced classes, the adaptation of SigLip~\cite{zhai2023sigmoid} is sufficient. The overall loss $L$ is computed as the sum of losses across all feature levels: $L = \sum_{i} L^i$. 


We use~\textbf{SigLip loss}~\cite{zhai2023sigmoid} instead of the original CLIP loss\cite{radford2021learning} because our architecture introduces two learnable text embeddings, each linked to multiple images. Specifically, image similarity is computed across both normal and abnormal prompts, while each prompt is compared against all images. This results in a similarity matrix capturing multiple valid associations rather than a strict diagonal mapping. While CLIP loss can compute image similarity across all texts, it cannot directly handle text similarity across multiple images, as each text embedding corresponds to multiple images in a batch. In contrast, SigLip loss processes image-text pairs independently, naturally supporting our many-to-one setup. This distinguishes our approach from previous work in medical AD~\cite{wang2022medclip,hua2024medicalclip,zhang2024mediclip,huang2024adapting}, with its contribution experimentally validated below.

\section{Experimental Analysis and Results}
\label{sec:experiments}
            We follow the latest SOTA: MVFA~\cite{huang2024adapting}, using a medical AD benchmark that spans five modalities and six datasets: brain MRI~\cite{baid2021rsna,bakas2017advancing,menze2014multimodal}, liver CT~\cite{bilic2023liver,landman2015miccai}, retinal OCT (composed of two datasets; OCT17~\cite{kermany2018identifying}, and RESC~\cite{hu2019automated}), chest X-ray (Chest)~\cite{wang2017chestx}, and digital histopathology (HIS)~\cite{bejnordi2017diagnostic}. BrainMRI, LiverCT, and RESC are used for both AC and AS, while OCT17, Chest, and HIS are relevant only for AC. We use the area under the Receiver Operating Characteristic curve (AUC), the standard medical AD metric, to report AUC for AC and AUC for AS. \\

        \noindent \textbf{Implementation Details.}
        We use the CLIP model with the ViT-L/14 architecture and 240-pixel input images, as in~\cite{huang2024adapting}. The model has 24 layers and the adapters were applied to the 6th, 12th, 18th, and 24th layers. Training is performed with the Adam optimizer with the learning rate of $1e^{-3}$, batch size 16, for 60 epochs. 
        Augmentation follows the strategy outlined in~\cite{huang2024adapting}. \\
    
\noindent \textbf{Comparisons with SOTA.}
        Table~\ref{tab:vanilla_comparison} presents the results of \method alongside SOTA. Methods labeled as unsupervised (referred to as Unsup) rely on large auxiliary datasets containing only normal samples, while \textit{few-shot} methods utilize a fixed set of 16 normal and abnormal samples.
        \method outperforms all SOTA across multiple datasets, except for OCT17~\cite{kermany2018identifying}, where April-GAN~\cite{chen2023zero} achieves better results. The slightly better performance of April-GAN~\cite{chen2023zero} on OCT17~\cite{kermany2018identifying} is likely due to the low inter-sample variability of the dataset, which is known to benefit memory bank-based methods that operate by comparing test samples with stored representations from the training set. On average, \method achieves best overall performance, surpassing the second-best method by 3.5\% in AC and 0.22\% in AS. The performance gain of \method can reach up to 25.79\% in AC and 13.23\% in AS.
        Table~\ref{tab:allshots} further compares few-shot methods for different numbers of normal/abnormal samples. On average, independent of the number of samples, \method performs the best, except for the OCT17~\cite{kermany2018identifying}, where all methods perform very similarly. Overall, as expected, the performance of the methods increases as more samples are added to the training data. Still, in the extreme case where only 2 normal samples and 2 abnormal samples are available, \method outperforms the others in 7 out of 9 cases.  \\

\begin{table}[t!]
\centering
\caption{
Comparisons with SOTA in terms of AUC (\%). Few-shot models use 16 samples per class. Best results are \textbf{bold}, second-best \underline{underlined}.}
\resizebox{\linewidth}{!}{
\begin{tabular}{l|l|l|c|c|c|cc|cc|cc|cc}
\hline
\textbf{} & \textbf{Method} & \textbf{Source} & \multicolumn{1}{c|}{\textbf{HIS}} & \multicolumn{1}{c|}{\textbf{Chest}} & \multicolumn{1}{c|}{\textbf{OCT17}} & \multicolumn{2}{c|}{\textbf{BrainMRI}} & \multicolumn{2}{c|}{\textbf{LiverCT}} & \multicolumn{2}{c|}{\textbf{RESC}} & \multicolumn{2}{c}{\textbf{Average}} \\ \cline{4-14}
& &  & \textbf{AC} & \textbf{AC}  & \textbf{AC}  & \textbf{AC} & \textbf{AS} & \textbf{AC} & \textbf{AS} & \textbf{AC} & \textbf{AS} & \textbf{AC} & \textbf{AS} \\ \hline

\multirow{4}{*}{Unsup}
 &CFLOWAD \cite{gudovskiy2022cflow} & WACV22 & 54.54 & 71.44 & 85.43 & 73.97 & 93.52 & 49.93 & 92.78 & 74.43 & 93.75 & 68.29 & 93.35 \\ 
  &RD4AD \cite{deng2022anomaly} & CVPR22 & 66.59 & 67.53 & {97.24} & 89.38 & 96.54 & 60.02 & 95.86 & 87.53 & 96.17 &  78.04 & 96.19\\ 
  &PatchCore \cite{roth2022towards} & CVPR22 & 69.34 & 75.17 & 98.56 & {91.55} & {96.97} & {60.40} & {96.58} & {91.50} &  {96.39} & 81.09 &  {96.65} \\ 
  &MKD \cite{salehi2021multiresolution} & CVPR22 &  {77.74} &  {81.99} & 96.62 & 81.38 & 89.54 & 60.39 & 96.14 & 88.97 & 86.60 &  {81.18} & 90.76\\ \hline

\multirow{6}{0.5cm}{Few shot}

 & DRA \cite{ding2022catching} & CVPR22 & 79.16 & {85.01} & \underline{99.87} & 82.99 & 80.45 & 80.89 & 93.00 & 94.88 & 84.01 & 87.13 & 85.82 \\
        
        & BGAD \cite{yao2023explicit} & CVPR23 & - & - & - & 88.05 & 95.29 & 78.79 & 99.25 & 91.29 & 97.07 & - & 97.20   \\
        
        & APRIL-GAN \cite{chen2023zero} & CVPRw23 & {81.16} & 78.62 & \textbf{99.93} & {94.03} & {96.17} & {82.94} & {99.64} & 95.96 & 98.47 & 88.77 & 98.09 \\

        & MediCLIP \cite{zhang2024mediclip} & MICCAI24 & 70.22 & 69.74 & 96.37 & 91.56 & 98.08 & 79.31 & 98.95 & 86.51 & 94.07 & 82.28 & 97.03\\
        
        & MVFA \cite{huang2024adapting} & CVPR24 & \underline{82.62} & \underline{85.72} & 99.66 & \underline{94.40} & \underline{97.70} & \underline{83.85} & \underline{99.73} & \underline{97.25} & \underline{99.07} & \underline{90.58} & \underline{98.83}\\
        
        & \method (Ours) &  & \textbf{90.14} & \textbf{88.15} & 99.71 & \textbf{95.9} & \textbf{97.97} & \textbf{91.46} & \textbf{99.74} & \textbf{99.11} & \textbf{99.45} & \textbf{94.08} & \textbf{99.05}\\ 
 \hline
\end{tabular}
}

\label{tab:vanilla_comparison}
\end{table}

        \begin{table}[t!]
\centering
 \caption{
 Comparisons with few-shot SOTA for $2$, $4$, and $8$ shots per class (AUC \%). Results for $16$ shots are in Table \ref{tab:vanilla_comparison}. Best results are \textbf{bold}, second-best \underline{underlined}.}
\resizebox{\linewidth}{!}{
\begin{tabular}{c|l|l|c|c|c|cc|cc|cc|cc}
\hline
\textbf{} & \textbf{Method} & \textbf{Source} & \multicolumn{1}{c|}{\textbf{HIS}} & \multicolumn{1}{c|}{\textbf{Chest}} & \multicolumn{1}{c|}{\textbf{OCT17}} & \multicolumn{2}{c|}{\textbf{BrainMRI}} & \multicolumn{2}{c|}{\textbf{LiverCT}} & \multicolumn{2}{c|}{\textbf{RESC}}   & \multicolumn{2}{c}{\textbf{Average}} \\ \cline{4-14}
 &  &  & \textbf{AC} & \textbf{AC}  & \textbf{AC}  & \textbf{AC} & \textbf{AS} & \textbf{AC} & \textbf{AS} & \textbf{AC} & \textbf{AS} & \textbf{AC} & \textbf{AS}\\ \hline

        \multirow{6}{*}{2} 
        & DRA \cite{ding2022catching} & CVPR22 & {72.91} & {72.22} & {98.08} & 71.78 & 72.09 & 57.17 & 63.13 & 85.69 & 65.59 & 76.3 & 66.93\\
        & BGAD \cite{yao2023explicit} & CVPR23 & - & - & -  & {78.70} & 92.42 & {72.27} & \underline{98.71} & 83.58 & 92.10 & -& 94.41 \\
      
        & APRIL-GAN \cite{chen2023zero} & CVPRw23 & 69.57 & 69.84 & \textbf{99.21}  & 78.45 & {94.02} & 57.80 & 95.87 & 
           {89.44} & {96.39} & 77.38 & 95.42\\

& MediCLIP \cite{zhang2024mediclip} & MICCAI24 & 64.49 & 61.69 & 93.4 & 85.13&\underline{ 97.39}&68.48& 97.09 & 83.96 & 96.01 & 76.2 & 96.83\\
        
        & MVFA \cite{huang2024adapting}& CVPR24 & \underline{82.61} & \underline{81.32} & 97.98 & \underline{92.72} & {96.55} & \underline{81.08} & {96.57} & \underline{91.36} & \textbf{98.11} & \underline{87.84} & \underline{97.07} \\

        & \method (Ours) &  & \textbf{83.62} & \textbf{84.56} & \underline{99.06} & \textbf{93.93} & \textbf{97.92}&\textbf{84.48}&  \textbf{99.39}&	\textbf{95.09} & \underline{97.18}& \textbf{90.12} & \textbf{98.16} \\
        \midrule
        
        \multirow{6}{*}{4} 
        & DRA \cite{ding2022catching} & CVPR22 & 68.73 & 75.81 & 99.06 & 80.62 & 74.77 & 59.64 & 71.79 & 90.90 & 77.28 & 79.12 & 74.61 \\
        
        & BGAD \cite{yao2023explicit} & CVPR23 & - & - & - & 83.56 & 92.68 & {72.48} & 98.88 & 86.22 & 93.84& - & 95.13 \\
        
        & APRIL-GAN \cite{chen2023zero} & CVPRw23 & {76.11}  & {77.43}  & \textbf{99.41} & {89.18} & {94.67} & 53.05 & 96.24 & 94.70 & {97.98} & 81.64 & 96.29 \\

& MediCLIP \cite{zhang2024mediclip} & MICCAI24 & 70.85 & 56.83 & 89.07 & 83.82 & 96.86 & \underline{81.53} & 98.61&	87.52&96.65& 78.27 & 97.37\\
        
        & MVFA \cite{huang2024adapting}& CVPR24 & \textbf{82.71}  & \underline{81.95} & \underline{99.38} & \underline{92.44} & \underline{97.30} & {81.18} & \textbf{99.73} & \underline{96.18} & \textbf{98.97} & \underline{88.97} & \underline{98.66} \\

        & \method (Ours) &  & \underline{80.05} & \textbf{88.10} &99.37 & \textbf{95.25}& \textbf{97.90} & \textbf{82.97} & \underline{99.29} & \textbf{96.62} & \underline{98.9}& \textbf{90.39}& \textbf{98.69}  \\

        \midrule
        \multirow{6}{*}{8} 
        & DRA \cite{ding2022catching} & CVPR22 & 74.33 & {82.70} & 99.13 & 85.94 & 75.32 & 72.53 & 81.78 & {93.06} & 83.07 & 84.61 &80.05 \\
        
        & BGAD \cite{yao2023explicit} & CVPR23 & - & - & - & 88.01 & 94.32 & {74.60} & {99.00} & 89.96 & 96.06 & - &96.46 \\
        
        & APRIL-GAN \cite{chen2023zero} & CVPRw23 & {81.70} & 73.69 & \textbf{99.75} & {88.41} & {95.50} & 62.38 & 97.56 & 91.36 & {97.36} & 82.88& 96.80 \\

& MediCLIP \cite{zhang2024mediclip} & MICCAI24 & 69.8 & 72.08 & 95.69 & 92.29 &\textbf{ 98.02} & \underline{86.32}	& 98.32	& 88.82	& 95.98&84.17 & 97.44\\

        & MVFA \cite{huang2024adapting}& CVPR24 & \underline{85.10} & \underline{83.89} & \underline{99.64} & \underline{92.61} & {97.21} & {85.90} & \underline{99.79} & \underline{96.57} & \textbf{99.00} & \underline{90.61} & \underline{98.66} \\

        & \method (Ours) &  & \textbf{87.45} &  \textbf{83.90} & 99.14 & \textbf{95.17} & \textbf{98.02} &\textbf{89.31}& \textbf{99.81} &\textbf{97.16}& \underline{98.85} & \textbf{92.02}& \textbf{98.89}\\
        
        
        
        


  \hline
\end{tabular}
}

 \label{tab:allshots}
\end{table}

        \noindent \textbf{Ablation studies.} 
        We conducted ablation studies on both the AC and AS tasks, reporting average results over three different seeds and six datasets to assess the overall effectiveness of \method. 
        \textbf{(a)} The impact of prompt design was evaluated by replacing our learnable prompt tokens (i.e., $[V_1][V_2]\ldots[V_M]$ and $[W_1][W_2]\ldots[W_M]$) with the hand-crafted templates used in~\cite{huang2024adapting,jeong2023winclip}. This substitution led to performance drops of 1.28\% for AC and 1.3\% for AS, highlighting the contribution of our learnable prompts. \textbf{(b)} Substituting the \texttt{[Objective]} term with ``medical image'' resulted in drops of 1.71\% in AC and 1.01\% in AS, further demonstrating the benefits of learnable prompts and context-specific information.
        We further examined our dual-branch design by performing two ablations. \textbf{(c)} removing one set of adapters (i.e., leaving 4 shared adapters for both normal and anomaly classes, instead of the 8 in \method) resulted in declines of 1.14\% for AC and 1.06\% for AS.
        \textbf{(d)} eliminating the signal from the opposite class while calculating \( S^i_n \) and \( S^i_{ab} \) (i.e., excluding the subtraction term from the calculation of the mentioned formulas), led to decreases of 1.41\% for AC and 1.24\% for AS.
        \textbf{(e)} We replaced SigLip Loss with the CLIP-based SoftMax loss, resulting in a performance drop of 1.48\% for AC and 1.1\% for AS. \textbf{(f)} The use of a single common head instead of separate heads, \( F_{\text{Det}} \) and \( F_{\text{Seg}} \), for the AC and AS tasks, respectively resulted in performance drops of 1.43\% for AC and 3.41\% for AS. To sum up, these ablation studies validate our design choices, demonstrating their positive contribution to both AC and AS tasks. \\
        


        \noindent \textbf{Cross-dataset analysis.} \method is compared with the best counterpart: MVFA~\cite{huang2024adapting} to assess generalization in a cross-dataset setting. Each model was trained on 16 samples from the datasets described above and tested on unseen target datasets of the same modality, as listed in Table \ref{tab:xvalidation}. As seen, \method consistently outperforms MVFA across all individual datasets, demonstrating superior cross-dataset generalization. Notably, \method achieves an average performance of 76.12\% compared to 74.53\% for MVFA.
             \begin{table}[t!]
\centering
\caption{Cross-dataset evaluation. AC is reported for all datasets. Best are \textbf{bold}.}
\resizebox{\linewidth}{!}{
\begin{tabular}{lc c c c c c}
\toprule
{Source} & \multicolumn{2}{c}{Chest} & \xspace \xspace BrainMRI \xspace \xspace  &  {\xspace \xspace OCT17 \xspace \xspace } & RESC \xspace \xspace  & \multirow{2}{*}{\xspace \xspace AVG} \\
\cmidrule(lr){2-3} \cmidrule(lr){4-4} \cmidrule(lr){5-6}
Target & NIHChest \cite{summers2019nih} & CheXpert \cite{irvin2019chexpert} & ADNI \cite{jack2008alzheimer} & \multicolumn{2}{c}{OCTDL \cite{kulyabin2024octdl}} &  \\ 
\midrule

MVFA \cite{huang2024adapting} 
& 61.91 & 80.41 & 53.94 & 88.47 & 87.94 & 74.53 \\

\method (Ours) & \textbf{62.44} & \textbf{81.99} & \textbf{57.35} & \textbf{90.74} & \textbf{88.09} & \textbf{76.12} \\

\bottomrule
\end{tabular}
}
\label{tab:xvalidation}
\end{table}


\section{Conclusion}
\label{sec:conclusion}

We introduced a few-shot AD architecture that leverages CLIP with multi-level adapters and prompt learning to model normal and abnormal classes separately. 
Our dual-objective strategy, formulated through subtraction, incorporates a contrastive effect by encouraging similarity within the same class and dissimilarity between opposing class. By integrating SigLIP loss, we further refine this separation process, as it can handle many-to-one relationship of images and learnable unpaired texts.
Extensive validation across diverse datasets demonstrates superior performance and strong generalization over SOTA methods, underscoring our approach's robustness for medical AD. However, a current limitation of \method is the assumption that learnable adapters and prompts are sufficient to bridge the gap between medical images and textual descriptions, while the modality gap remains unaddressed explicitly. Future work will explore the method's zero-shot potential and extend it to multi-modal AD, handling diverse training modalities and unseen anatomical regions simultaneously.


\begin{credits}
\subsubsection{\ackname} This preprint has not undergone peer review or any post-submission revisions or corrections. The Version of Record of this contribution has been accepted to MICCAI 2025; the DOI can be found in arXiv comment section.
We acknowledge the financial support  of the PNRR project FAIR - Future AI Research (PE00000013),  
under the NRRP MUR program funded by the NextGenerationEU.

\end{credits}
%
%
%
\bibliographystyle{splncs04}
\bibliography{biblio}

\end{document}